\documentclass[sigconf]{acmart}
\usepackage{tabu}
\usepackage{booktabs} 
\usepackage{multirow}
\usepackage{booktabs, caption, makecell}
\usepackage{amsmath,amsthm,amssymb}
\usepackage{algorithm,algorithmic}

\copyrightyear{2019}
\acmYear{2019} 
\setcopyright{iw3c2w3}  
\acmConference[WWW '19]{Proceedings of the 2019 World Wide Web Conference}{May 13--17, 2019}{San Francisco, CA, USA}
\acmBooktitle{Proceedings of the 2019 World Wide Web Conference (WWW '19), May 13--17, 2019, San Francisco, CA, USA}
\acmPrice{}
\acmDOI{10.1145/3308558.3313520}
\acmISBN{978-1-4503-6674-8/19/05}

\usepackage{balance}

\begin{document}

\title{Sampled in Pairs and Driven by Text: A New Graph Embedding Framework}


\author{Liheng Chen}
\affiliation{
\institution{Shanghai Jiao Tong University}
}
\email{lhchen@apex.sjtu.edu.cn}

\author{Yanru Qu}
\affiliation{
	\institution{Shanghai Jiao Tong University}
}
\email{kevinqu@apex.sjtu.edu.cn}

\author{Zhenghui Wang} 
\affiliation{
	\institution{Shanghai Jiao Tong University}
}
\email{felixwzh@apex.sjtu.edu.cn}

\author{Weinan Zhang}
\affiliation{
	\institution{Shanghai Jiao Tong University}
}
\email{wnzhang@apex.sjtu.edu.cn}

\author{Ken Chen}
\affiliation{
	\institution{Synyi LLC.}
}
\email{chen.ken@synyi.com}

\author{Shaodian Zhang}
\affiliation{
	\institution{Synyi LLC.}
}
\email{shaodian@apex.sjtu.edu.cn}

\author{Yong Yu}
\affiliation{
	\institution{Shanghai Jiao Tong University}
}
\email{yyu@apex.sjtu.edu.cn}

 \renewcommand{\shortauthors}{Liheng Chen et al.}
\newcommand{\acqu}[1]{#1}
\newcommand{\clh}[1]{{\color{purple}#1}}
\newcommand{\qu}[1]{{\color{blue} #1}}
\newcommand{\zh}[1]{{\bf \color{purple} [wang says ``#1'']}}
\newcommand{\zwn}[1]{{\bf \color{red} [weinan says ``#1'']}}
\newcommand{\mN}{\mathcal{N}}
\newcommand{\be}{\mathbf{e}}
\newcommand{\rae}{\overrightarrow{\be}}
\newcommand{\lae}{\overleftarrow{\be}}
\newcommand{\Be}{\mathbf{e}}
\newcommand{\todo}[1]{\textbf{``#1''}}

\begin{abstract}
In graphs with rich texts, incorporating textual information with structural information would benefit constructing expressive graph embeddings. 
\acqu{Among various graph embedding models, random walk (RW)-based is one of the most popular and successful groups.
	However, it is challenged by two issues when applied on graphs with rich texts:}
(i) \textit{sampling efficiency}: deriving from the training objective of RW-based models (e.g., DeepWalk and node2vec), we show that \acqu{RW-based models are likely to generate large amounts of redundant training samples due to three main drawbacks.}
(ii) \textit{text utilization}: these models have difficulty in dealing with zero-shot scenarios where graph embedding models have to infer graph structures directly from texts.
To solve these problems, we propose a novel framework, namely Text-driven Graph Embedding with Pairs Sampling (TGE-PS).
TGE-PS uses Pairs Sampling (PS) \acqu{to improve the sampling strategy of RW, being able to reduce }
$\sim$99\% training samples \acqu{while preserving competitive performance.} 
TGE-PS uses Text-driven Graph Embedding (TGE), \acqu{an inductive graph embedding approach, to generate node embeddings from texts.} 
\acqu{Since each node contains rich texts, TGE is able to generate high-quality embeddings and provide reasonable predictions on existence of links to unseen nodes.}
We evaluate TGE-PS on several real-world datasets, and experiment results demonstrate that TGE-PS produces state-of-the-art results 
on both traditional and zero-shot link prediction tasks.
\end{abstract}

%
%




\maketitle

\section{Introduction}

Graph provides a fundamental tool to represent interconnected entities \acqu{(e.g., articles, diseases)}
and their attributes \acqu{(e.g., entity description)}.
Graphs with rich text information are ubiquitous in many fields \cite{hamilton2017representation}, 
and there is often a strong dependency between graph structure and text structure in these graphs.
Textural information may also expose structural information \cite{zhang2018network}.
Hence, it is \acqu{promising} 
to better utilize textual information in graphs.

Graph embedding is famous for its efficient representations for entities in graphs~\cite{goyal2018graph,nishana2013graph}.
A series of models are proposed to maximize edge reconstruction probability with different proximities ~\cite{cai2018comprehensive}, e.g., LINE~\cite{Tang2015}, DeepWalk~\cite{Perozzi2014} and node2vec~\cite{Grover2016}.
Although these models \acqu{are widely used in mapping nodes to low-dimensional dense vectors,} 
\acqu{they are not well designed for graphs with rich text information.}
To solve this problem, models like DDRW \cite{li2016discriminative}, GENE \cite{chen2016incorporate}, PPNE \cite{li2017ppne} and Tri-DNR \cite{pan2016tri} focus on preserving vertex labels.
Especially, TADW~\cite{yang2015network}, CANE~\cite{Tu2017} and Paper2Vec~\cite{ganguly2017paper2vec} are proposed to utilize textural information \acqu{with} 
effectiveness in many scenarios. 

However, these graph embedding models still need to resolve two issues.
The first issue is \textit{sampling efficiency}.
The optimization goals of these models can be \acqu{summarized} as maximizing pairwise node similarity,
\acqu{thus the number of training node pairs is critical to training time and can be used as a metric of sampling efficiency.} 
\acqu{These models usually use edges as training node pairs directly like LINE, or sample training sequences using random walk (RW) and extract node pairs within a specific shortest distance like DeepWalk and node2vec.}
RW generalizes the idea of directly sampling edges and shows 
better experimental performance under similar settings. 
However, our theoretical and empirical analyses show that RW samples redundant node pairs which severely lags efficiency.  

The second issue is \textit{text utilization}.
Text utilization models like TADW and CANE mainly rely on training node embeddings  (NE) and text embeddings  (TE) together (denoted as NE+TE).
In zero-shot settings, where link existence with previously unseen nodes is to be predicted and the only reliable information is texts attributed to nodes
, humans can make inference by assuming textual connections even if he has little prior knowledge about this field.
However, NE+TE is incapable of dealing with this scenario due to two drawbacks: (i) it requires both NE and TE to generate high-quality representations, but NE is missing for unseen nodes, and (ii) even if it uses only TE part to predict, empirical results show that it may be no better than simple text matching. 
This reveals that NE+TE does not utilize text information sufficiently.

In this paper, we propose a novel Text-driven Graph Embedding with Pairs Sampling (TGE-PS) framework.
TGE-PS \acqu{uses Pairs Sampling (PS) to efficiently generate training samples, and uses Text-driven Graph Embedding (TGE) to produce final node representations.}
We theoretically analyze the redundant sample phenomenon of RW from three perspectives
, and propose PS to improve sampling efficiency.
PS samples center-neighbor node pairs directly from central node's neighborhood.
Our experiment results on 6 datasets show that PS produces competitive or even better results in link prediction task with much fewer training samples (saving $\sim$99\% samples) compared with RW. 
In embedding stage, we propose inductive TGE method.
TGE incorporates text information with structure information and encodes character- and word-level embeddings into node embeddings while following graph structure.
\acqu{Since node embeddings are generated from text embedding in TGE, they can be applied to zero-shot scenarios.}
The comparison between TGE-PS and other strong baseline models shows that TGE-PS
produces remarkably good results in traditional and zero-shot link prediction tasks.

\section{Related Works}

Recent years have witnessed various graph embedding models with applications in link prediction\cite{lu2011link,backstrom2011supervised, liben2007link}, node classification\cite{yang2011like, tsoumakas2007multi,sen2008collective,kazienko2012label}, clustering\cite{belkin2002laplacian,fortunato2010community}, 
recommendation \cite{zhang2017regions,xie2016learning,wang2018ensemble},
knowledge graph \cite{bordes2011learning}, etc. 
These graph embedding models can be categorized into three classes, factorization-based, random walk (RW)-based, and deep learning-based\cite{goyal2018graph, cai2018comprehensive}.
Factorization-based models focus on the connections among nodes and using matrix factorization (MF) to learn the low-rank representations of nodes \cite{ahmed2013distributed, cao2015grarep, ou2016asymmetric, yang2015network}.
TADW \cite{yang2015network} uses MF to decompose the transition matrix with textual information incorporated.
RW-based models explore the neighborhood of each node through sampling paths, and thus can maintain local structural information in node embeddings.
Among RW-based models, DeepWalk \cite{Perozzi2014} and node2vec \cite{Grover2016} are the most representative, and DeepWalk can be regarded as a special case of node2vec.
Deep learning-based models mainly use deep representation learning techniques to improve the quality of node embeddings \cite{wang2016structural, cao2016deep, Tu2017,jia2019communitygan}.

There are also models that do not belong to these three classes.
LINE \cite{Tang2015} proposes to train graph embeddings through first-order and second-order proximities and is efficient in large-scale networks. 
Besides, LINE claims to preserve both local and global network structures.
GraphGAN \cite{wang2018graphgan} adopts GAN \cite{goodfellow2014generative} framework to learn the underlying true connectivity distribution implicitly.
The generator produces ``fake edges'' while the discriminator tries to tell generated node pairs from ground truth.
It also proposes Graph Softmax to boost its efficiency in training.

Among graph embedding models, RW-based models present robust and remarkable performance on various datasets.
The training of RW-based models have two phases, the sampling phase and the optimization phase.
Different sampling policies are proposed to generate high-quality node sequences to explore the neighborhoods of certain nodes.
Especially, DeepWalk adopts the simplest policy where each node is generated only depending on its predecessor, while node2vec adopts a biased policy to trade off between Breadth-first Sampling (BFS) and Depth-first Sampling (DFS).
After obtaining sufficient node sequences, RW-based models use SkipGram \cite{mikolov2013efficient, mikolov2013distributed} model to maximize the log-probability of observing a network neighborhood for a node conditioned on its feature representation.
However, we concern the sampling strategies of RW-based models have intrinsic drawbacks, which we discuss in Section~\ref{Sec-Random Walk}. 

For graphs with rich texts, many models are proposed to incorporate textual information with structural information.
TADW \cite{yang2015network} 
incorporates text features  under the framework of matrix factorization.
CANE \cite{Tu2017} uses convolutional neural networks and mutual attention mechanism to learn text embeddings, which are interacted with node embeddings via vector inner product.
These models encode structural information and textual information into two separate embedding spaces and generate final node representations from the interactions between these two spaces.
Paper2Vec~\cite{ganguly2017paper2vec} pre-trains node embeddings \acqu{with text embeddings in Skip-gram model, and then the node embeddings are trained with node2vec.} 
STNE~\cite{liu2018content} ``self-translates'' sequences of text embeddings into sequences of node embeddings. 
All the above models rely on known connections to generate graph embeddings, thus are not applicable to zero-shot scenarios.
Although some previous works \cite{xie2016representation,fu2015transductive,kipf2016semi} discuss graph representation learning in zero-shot setting, they do not solve the same problem as we do.

When processing texts in zero-shot scenarios, it is common to encounter out-of-vocabulary words.
In NLP field, character-level embeddings have proven success like Neural Machine Translation \cite{chung2016character, lee2017fully, ling2015character} and Named Entity Recognition \cite{ma2016end, santos2015boosting}. 
These models are based on a hybrid of character- and word-level embeddings, thus can automatically capture patterns in the sub-word level and present advantages in dealing with unseen words. 
\begin{figure*}[htbp]
	\centering
	\vspace{-0.3cm}
	\includegraphics[width=0.85\textwidth]{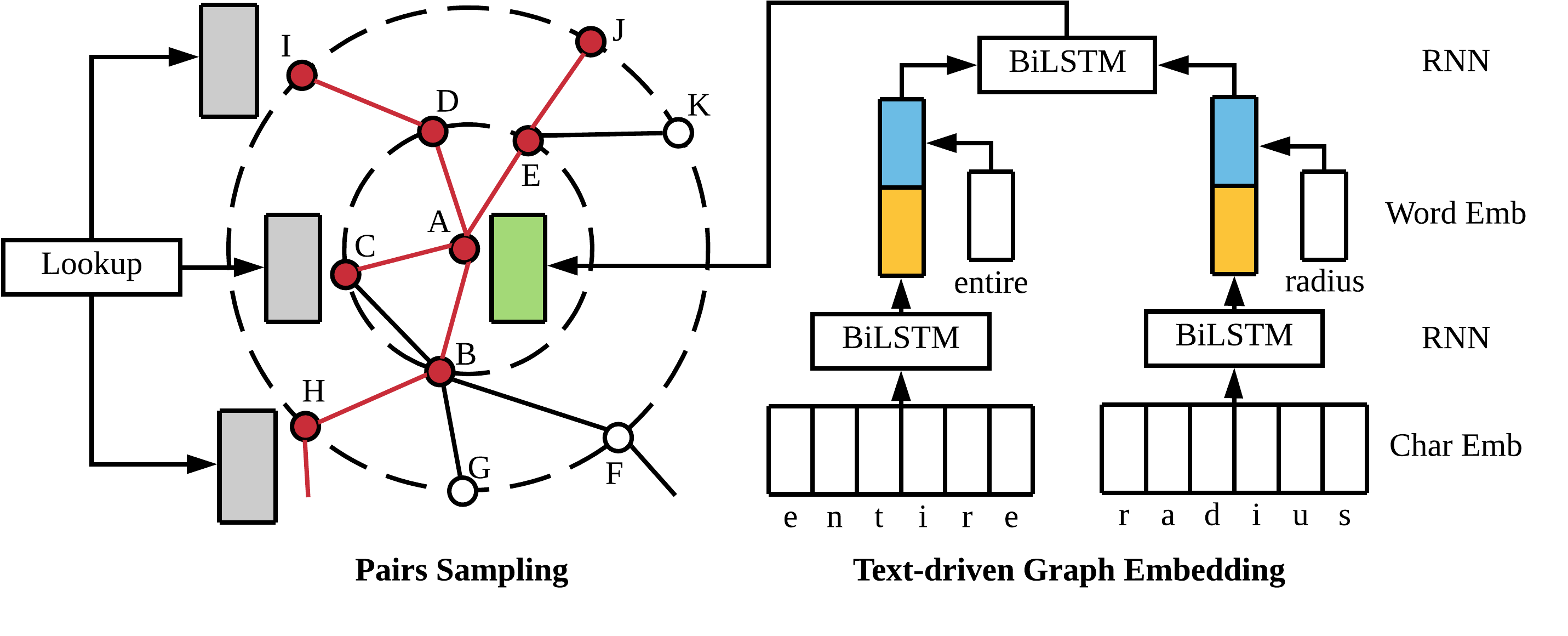}
	\vspace{-0.6cm}
	\caption{Model architecture of TGE-PS.
		\emph{Note:} Circles represent nodes and  colored boxes represent embeddings. Blue boxes represent word-level embeddings, yellow boxes represent character-based word embeddings, green boxes represent central embeddings and gray boxes represent neighbor embeddings. 
		The sampled nodes and paths are colored as red.
	}
	\label{Fig-Model Architecture}
	\vspace{-0.5cm}
\end{figure*}

\section{Text-driven Graph Embedding with Pairs Sampling Framework}
In this section, we firstly give definitions of notions, and then introduce PS and TGE separately.
We show the architecture of TGE-PS in Fig.~\ref{Fig-Model Architecture}, with fully-connected and lookup layers omitted.

\subsection{Definitions}	
Let $G =   (V, E)$ be the given graph and $f:V\to \mathbb{R}^n$ be the mapping function from the node set to the $n$-dimensional embedding space.
We denote the central and context embeddings of node $v_i$ as $\Be_i=f  (v_i)$ and $\Be'_i=f' (v_i)$, and the trainable parameters as $\theta$. We define the distance  $\mathcal{L}(u, v)$ between $u$ and $v$ as the length of the shortest path between them. 
And we define node $v_i$'s $o$-neighborhood $\mN_i^o$ 
as a set of nodes within a given distance $o$ from $v_i$, $\mathcal{N}^o  (v_i) = \{v_j \in V | \text{dist}  (v_i, v_j) \leq o\}$. 
For every node $v_i$, we call one node $v_j^o$ as an $o$-th order neighbor of $v_i$ when the distance between $v_i$ and $v_j$ is $o$.
\subsection{Pairs Sampling Method}
\subsubsection{Revisit Random Walk}
Since DeepWalk can be regarded as a special case of node2vec \cite{Perozzi2014,Grover2016}, our discussion mainly focuses on node2vec. 
The objective of node2vec is to maximize the log-probability of observing the $o$-neighborhood $\mN^o_i$ of a node $v_i$ 
as in Eq. \eqref{Eqn-Original Objective}, and following  conditional independence assumption of SkipGram, node2vec changes its objective to maximize the log-probability of all center-neighbor node pairs as in Eq. \eqref{Eqn-Objective}.
Following symmetry in feature space assumption, the conditional probability of a node pair is defined as in Eq. \eqref{eq:pr}, where $\tau_{i,j}$ is the abbreviation of the score function $\tau(v_i,v_j)$.
It is worth noting that, the score function $\tau_{i,j}$ is asymmetric since the input $v_j$ is conditioned on the other input $v_i$.
With Eq.~\eqref{Eqn-Objective} and Eq.~\eqref{eq:pr}, the objective becomes Eq. \eqref{eq:score}, where $Z_i$ denotes the normalizing term, and each window has size $2k$.
$Z_i$ is usually approximated by hierarchical softmax or negative sampling in training. 
And we mainly focus on the scoring terms $\sum_i{\sum_{j}{\tau_{i,j}}}$.

\begin{eqnarray}
&&\max_{\theta} \sum_{v_i \in V}{\log Pr(\mN^o_i|v_i)} \label{Eqn-Original Objective} \\
&= & \max_{\theta} \sum_{v_i \in V}{\sum_{v_j\in \mN^k_i}{\log Pr(v_j|v_i)}} \label{Eqn-Objective} \\
&= & \max_\theta \sum_{v_i \in V}{(\sum_{v_j\in \mN^k_i}{\tau_{i,j}}) - |\mN^k_i|\log Z_i} \label{eq:score}
\end{eqnarray}
\begin{equation} \label{eq:pr}
Pr(v_{j}|v_i) = \frac{\exp{(\tau_{i,j})}}{\sum_{v_k\in V}\exp{(\tau_{i,k})}}
\end{equation}

From the training perspective, node2vec defines the inner product of a central embedding and a context embedding to represent the score function, i.e., $\mathbf{e}_j'^\top \mathbf{e}_i$.
The training objective of the scoring terms becomes Eq. \eqref{eq:opt},
where $s$ denote a node sequence, $S$ is the set of all sequences, $\omega_i^s$ is the abbreviation of the window function $\omega(v_i,s)$ 
which denotes the nodes within the window of $v_i$ when $v_i$ appears in $s$
and $j'$ denotes the position of $v_j$ in $s$.
We transform Eq.~\eqref{eq:opt} to 
Eq.~\eqref{eq:obj1} and Eq.~\eqref{eq:obj2}, 
where $T$ denotes the sampling time of RW starting from $v_i$, $T_i$ denotes the amount of sequences where $v_i$ is not the starting point, $T_{j|i}$ denotes the amount of $v_j$ appearing in $N_i^k$ when $v_i$ appears in sequence $s$.
It is worth noting that, each node serves as the starting point in $T$ sequences, and each node also appears in the sequences starting from the other nodes.
Thus, there are $T + T_i$ windows centering at $v_i$ to be optimized in the sequences.
$\alpha_i$ is the ratio of $v_i$ being more ``important'' than the other nodes, e.g., bridge nodes are likely to be sampled more frequently, and these nodes have larger $\alpha$.
$\beta_{j|i}$ is the probability of $v_j$ sampled in the neighborhood $\mathcal{N}_i^o$ of $v_i$. 
Since there are $T+T_i$ windows centering at $v_i$, there are $2k(T + T_i)$ center-neighbor node pairs.
Thus $2k(T+T_i) = \sum_{v_j \in N_i^k} T_{j|i}$, and $\beta_{\cdot|i}$ reflects the distribution of the neighbor nodes appearing in training samples.
Till now, we get the ideal training objective of RW regardless of the sampling strategies.
In another word, any RW policies will converge to Eq.~\eqref{eq:obj2} when the amount of sampled sequences approaches infinity.

\begin{eqnarray}
&&\max_{\theta}\sum_{v_i \in V} \sum_{s\in S} \sum_{v_{j'} \in \omega_i^s} \Be_{j'}'^\top \Be_i \label{eq:opt}\\
&=&\max_{\theta}\sum_{v_i \in V} (T + T_i) (\sum_{v_j\in \mN^k_i} \frac{T_{j|i}}{2k(T + T_i)} \Be_{j'}'^\top \Be_i) \label{eq:obj1}\\
&=&\max_{\theta}T \sum_{v_i \in V} (1 + \alpha_i) (\sum_{v_j \in \mN^k_i} \beta_{j|i} \Be_{j'}'^\top \Be_i) \label{eq:obj2}
\end{eqnarray}

\subsubsection{Inefficiency of Random Walk}\label{Sec-Random Walk}
We introduce the sampling strategy of node2vec at first.
For every node $v_i$ in the graph, node2vec simulates a random walk with length $L$. During the sampling process, given previous node $t$ and current node $v$, the next node $x$ is sampled from the following distribution:

\begin{equation}
Pr(x|v)=\begin{cases}
\frac{p_{vx}}{Z} & \text{ if } (v, x)\in E \\
0 & \text{ otherwise}
\end{cases} \nonumber
\end{equation}
where $p_{vx}$ is the unnormalized transition probability from nodes $v$ to $x$, and $Z$ is the normalizing term. 
The unnormalized transition probability $p_{vx}$ is set to be $\alpha(t, x; p, q)\cdot w_{vx}$, where
$\alpha(t, x; p, q)$ is a function of $\text{dist}(t,x)$,
$w_{vx}$ denotes the edge weight of $(v, x)$, $\text{dist}(t,x)$ denotes the shortest path length between $t$ and $x$,
$p$ and $q$ control the searching strategy of random walks. 
Without loss of generality, we take $w_{vx} = 1$.

DeepWalk can be regarded as a special case of node2vec when $p=q=1$.
TADW has proved that DeepWalk is reconstructing the transition matrix $M$.
Induced from this conclusion, we view node2vec as reconstructing a biased transition matrix $M'$.
Thus in DeepWalk, $\beta_{j|i}$ in Eq.~\eqref{eq:obj2} are actually the non-zero elements of the $i$-th row of $M$ in DeepWalk, or the $i$-th row of $M'$ in node2vec.
The sampling strategy of node2vec is an exploration of neighborhoods with second-order Markov Property.
We analyze the inefficiency of RW in 3 aspects. 

\textbf{Biased Objective.}
Comparing Eq.~\eqref{eq:score} and ~\eqref{eq:obj2}, we find node2vec is biased from its log-probability objective since the neighborhoods $N_i^k$ are weighted by different $\alpha_i$.
$\alpha_i$ can be regarded as an ``importance ratio'', given the previously mentioned bridge nodes example.
Hence node2vec introduces a prior distribution implicitly, making its training objective biased from its proposal.
If we follow Eq.~\eqref{Eqn-Original Objective} strictly, $\alpha$ should be all zero and we should only sample each neighborhood $T$ time, which can reduce a lot of training samples.

\textbf{Interconnected Neighbors.}
The target of node embedding is to encode structural information in an embedding space.
It is a free lunch to assume when optimize a central node $v_i$, its neighbor nodes $v_j \in N_i^k$ have already been trained and the context embedding $\Be_j'$ have already been encoded with the structural information of $v_j$ respectively 
after training RW for some time, 
otherwise, the target of node embedding becomes ill-posed and can never be achieved.
Thus center-neighbor connections are much more important than neighbor-neighbor connections, and the neighbor-neighbor connections within one neighborhood will become the center-neighbor connections of other neighborhoods.
With these observations, we concern RW is less efficient since it results in lots of neighbor-neighbor connections when training central nodes.
Therefore, adopting a stronger assumption that alleviates variance of distribution $\beta_{\cdot|i}$ is still possible to produce good results while decreases sampling complexity.

\textbf{Revisiting Nodes.}
DeepWalk has the first-order Markov property since it only remembers the current node.
node2vec is second-order Markov since it remembers the current node and the previous node.
When the window size is larger than the Markov order, RW cannot prevent from revisiting a previously visited neighbor node.
The worst case is revisiting the central node itself, because that introduces self-loops and doubles the training time of the central node.
In ideal cases, revisiting is equivalent to approximate the transition matrix, where $\beta_{j|i} = M_{i,j}, \tau_{i,j} = \beta_{j|i} \Be_j'^\top \Be_i$.
Since $\beta_{j|i}$ is a constant determined by the graph structure, the node embeddings $\Be_i$ and $\Be_j'$ are approximating $\tau_{i,j} / \beta_{j|i}$, where $\tau_{i,j}$ is the unormalized probability.
The ideal sample complexity of a neighborhood $\mathcal{N}_i^k$ is determined by the minimal sampling probability $\beta_{j*|i}$, by assuming every node in the neighborhood $\mathcal{N}_i^k$ gets sufficient training samples.
For simplicity, we say the sample complexity of $\mathcal{N}_i^k$ is $O(1/\beta_{j*|i})$.
And this complexity must be greater than $O(|\mathcal{N}_i^k|)$ since $\beta_{j*|i} \le 1/|\mathcal{N}_i^k|$.
This means any distribution $\beta$ over $\mathcal{N}_i^k$ results in a higher sample complexity than uniform distribution $\beta_{\cdot|i} = 1/|\mathcal{N}_i^k|$.

\subsubsection{Method Introduction}\label{Sec-Method Introduction}
The intuition of Pairs Sampling (PS) has three points: as for the \textbf{biased objective} problem, it is desirable to sample each neighborhood with same time; as for the \textbf{interconnected neighbors} problem, it is desirable to sample center-neighbor pairs directly from a neighborhood; as for the \textbf{revisiting nodes} problem, it is promising to introduce higher-order Markov property.

To obtain the training node pairs set $\mathcal{P}$, we sample the neighbor nodes in the neighborhood of $v_i$ in the following process:

\begin{itemize}
\item[1.] Add all first-order neighbor nodes of $v_i$ into $\mathcal{N}^O_i$.
\item[2.] If $O>1$, for every $v^o_j$ ($1\leq o\leq {O-1}$), sample a next-order neighbor node $v^{o+1}_k$ from the following distribution: 
\begin{equation}
Pr(v^{o+1}_k|v^o_j) = \begin{cases}
\frac{1}{Z_j} & \text{ if }   (v^o_j, v^{o+1}_k) \in E\text{ and } \mathcal{L}(v_i, v^{o+1}_k) = o+1 \\
0 & \text{ otherwise} .
\end{cases} \nonumber
\end{equation}
where $Z_j$ is the number of $o+1$-order neighbors connected to $v_j^o$. The sampled $v^{o+1}_k$ is added into $\mathcal{N}^O_i$.
\item[3.] For each neighbor node $v^o_j$, add node pair $(v_i, v^o_j)$ into pairs set $\mathcal{P}$.
\item[4.] Repeat for $N$ times.
\end{itemize}

We take the graph in Fig. \ref{Fig-Model Architecture} as an example to illustrate how PS works. 
In this graph, $A$ is the central node, with dashed circles indicating the neighbors of the same order. 
We color sampled neighbors as red circles and leave the others as white ones. 
Following the aforementioned process, first-order neighbors $\{B, C, D, E\}$ are all sampled. 
$H$ is sampled from $\{F, G, H\}$ as the successor of $B$.
$I$ is sampled as the successor of $D$.
$J$ is sampled from $\{J, K\}$ as the successor of $E$.
Unsampled nodes are ignored in the next iteration of sampling, only sampled nodes in this order can be used to generate the next-order samples, e.g., $H$ continues searching while $F$ stops. 
By now PS samples a set of node pairs $\{(X,A)\}|X\in \{B,C,D,E,H,I,J\}\}$.

By restricting the max order of neighbors and the number of successors for each node to be at most 1, we successfully set an upper bound for the total number of node pairs. 
In fact, under the same training objective, pairs sampled by different sampling strategies will converge to different distributions of $\alpha$ and $\beta$. 
Theoretical optimal distributions are so far too complex to compute, but we show empirically that pairs sampled by Pairs Sampling can yield competitive results as RW-based does.
\subsection{Text-driven Embedding Model}

\subsubsection{Intuition}

Previous graph embedding models focus on generating graph embeddings from only structural information (denoted as NE) or incorporating text attributes with structural information (denoted as NE+TE). 
NE+TE models can be regarded as encoding structural information into an NE space and encoding textual information into a TE space. 
Even though complicated interactions are explored between these two spaces, 
we concern
the textual information has not been fully utilized, 
and propose to generate graph embeddings from text embeddings (denoted as TE2NE). 
TE2NE is more suitable for large graphs with strong text dependency. 
A large-scale graph with rich texts may contain million- to billion-level nodes, where the number of nodes is much larger than the number of words.
Besides, TE2NE can also apply to zero-shot scenarios, since no explicit NE is required in inference. 
Therefore, we propose the Text-driven Graph Embedding (TGE) method that makes the most of textual information by projecting textual information into the NE space.
To model the text, we adopt bidirectional LSTM (BiLSTM) \cite{ma2016end} for its success in Nature Language Processing field \cite{mikolov2012context,mikolov2010recurrent}. 
To deal with out-of-vocabulary words in unseen nodes, we adopt character-level embeddings.
Character-level embeddings have proven success in NLP tasks~\cite{chung2016character,ma2016end}.
The advantages of character-level embeddings over word-level embeddings are summarized by~\cite{chung2016character}.
Hence, we adopt character-level embeddings in addition to word-level embeddings. 

\subsubsection{Generating Embeddings}

We denote the set of words as $\mathcal{D}^w$, the set of characters as $\mathcal{D}^c$ and text of node $v_i$ to be $t_i = \{w_{ij}, 1\leq j\leq |t_i|\}$, where $w_{ij} \in \mathcal{D}^w$ and $|t_i|$ is the length of $t_i$. Each word $w_{ij}$ contains characters as $w_{ij}=\{c_{ijk}, 1\leq k\leq |w_{ij}|\}$, where $c_{ijk} \in \mathcal{D}^c$ and $|w_{ij}|$ is the length of $w_{ij}$. We denote character- and word-level embedding vectors as $\be^c$ and $\be^w$ respectively.

We start by generating embedding of node $v_i$. We feed the sequence of character embeddings $\be^c_{ij, 1:|w_{ij}|}$ into character-level BiLSTM and obtain a character-based word embedding $\be^{w'}_{ij}$ as in Eq. \eqref{eq:char emb}.
The character-based word embedding is concatenated with the corresponding word embedding $\textbf{e}^w_{ij}$ and fed into the word-level BiLSTM layer as in Eq. \eqref{eq:word emb}.

\begin{eqnarray}
\be^{w'}_{ij} & = & \text{BiLSTM}^c(\be^c_{{ij}, 1:|w_{ij}|})\label{eq:char emb} \\
\textbf{e}_i & = & \tanh(W{\text{BiLSTM}}^w([\be^w_{i, 1:|T_i|}; \be^{w'}_{i, 1:|T_i|}])+b) \label{eq:word emb}
\end{eqnarray}

Now we obtain the text-based node embedding $\be_i$ of node $v_i$. We also set up a lookup layer which embeds $v_i$ into $dim$-dimensional structure-based vector $\be'_i$ that is used to help train $\be_i$, and outputs $\be_i$ as the embedding vector of $v_i$. 

\subsubsection{Training}

The training objective is Eq.~\eqref{Eqn-Original Objective}. 
To reduce computation complexity, we 
define the loss function to be

\begin{equation}
\mathcal{L}_\text{sim}(v_i, v_j) = -\log(\sigma(\textbf{e}'^\top_j \textbf{e}_i)) - \sum_{v_k\sim P(v)}^{N_\text{neg}}{\log(\sigma(-\textbf{e}'^\top_k \textbf{e}_i))} \nonumber
\end{equation}

where $\be_i$, $\be'_j$ and $\be'_k$ denotes embeddings of the central node, the neighbor node in a pair, and the randomly sampled negative node, $N_\text{neg}$ denotes the number of negative samples \cite{mikolov2013distributed}, $\sigma$ denotes the sigmoid function and $P(v)\propto d_v^{3/4}$ denotes the distribution of nodes when sampling negative samples, where $d_v$ is the degree of $v$. 
We also apply $L_2$-regularization on parameters and embeddings
We use AdaGrad optimizer \cite{duchi2011adaptive} to minimize the loss.

\section{Experiments}
\subsection{Datasets}\label{Sec-Datasets}

To verify the effectiveness and efficiency of PS, we conduct a series of experiments over 6 datasets. 
We list their details in Tab.~\ref{Tab-Details of Datasets}, where $|V|$ and $|E|$ refer to number of nodes and edges, respectively. 
In practice, we keep all nodes and $p\%$ edges of the dataset $Data$ for training and denote this setting as $Data@p\%$. 
Before conducting our experiments, we pre-process texts including lowering all characters, removing stop words and discarding punctuations.

\begin{table}[htbp]
	\centering
	\vspace{-0.3cm}
	\caption{Details of Datasets}
	\vspace{-0.3cm}
	\begin{tabular}{c|ccc}
		\hline
		\textbf{Datasets} & $|V|$ & $|E|$ & Type  \\
		\hline
		Cora \cite{mccallum2000automating} & 2,211 & 5,214 & Citation Graph \\
		Facebook \cite{snapnets} & 4,039 & 88,234 & Social Network \\
		Zhihu \cite{Tu2017} & 10,000 & 43,894 & Q\&A Datasets \\
		AstroPh \cite{snapnets} & 18,772 & 198,110 & Co-work Network \\
		HepTh \cite{snapnets}  & 27,400 & 352,542 & Citation Graph \\
		SNOMED \cite{donnelly2006snomed,millar2016need} &391,892 & 2,047,749 & Health Terminology \\
		\hline
	\end{tabular}
	\label{Tab-Details of Datasets}
	\vspace{-0.3cm}
\end{table}

\subsection{Baselines} \label{Sec-Baselines}

We evaluate our model against several graph embedding models:
\textbf{LINE} \cite{Tang2015}, \textbf{DeepWalk} \cite{Perozzi2014} \textbf{node2vec} \cite{Grover2016}, \textbf{CANE} \cite{Tu2017}, \textbf{Paper2Vec} \cite{ganguly2017paper2vec} and \textbf{STNE} \cite{liu2018content}.
Since all listed models
are not capable of fitting in zero-shot scenarios. Therefore, we design a rule-based embedding method, \textbf{Text Matching}, which represents each node as the average vector of pre-trained word embeddings from Glove~\cite{pennington2014glove} for every word in the text.
The performance of Text Matching reflects the dependency between graph and text structure.

\subsection{Evaluation Metrics}
In evaluation, we sample an unobserved link $(v_c, v_p)$ and a non-exist link $(v_c, v_n)$ as the positive and negative samples for each node $v_c$, respectively. we adopt AUC (Area Under Curve) \cite{hanley1982meaning} as the metric in two different ways:
we use the a portion of pairs to train a logistic regression classifier, use the trained classifier to infer the connectivity of another set of pairs and compute the final AUC score as \textbf{AUC$_\text{LR}$};
we directly compute the times of $\be^\top_c \be_p > \be^\top_c \be_n$ as \textbf{AUC$_\text{pair}$}, similar to that in \cite{lu2011link}. 
We report the results of all experiments in two columns, which represent AUC$_\text{LR}$ and AUC$_\text{pair}$ respectively.

\subsection{Experiments of Pairs Sampling} \label{Sec-Comparison between Random Walk and Pairs Sampling}

\subsubsection{Theoretical Analysis}

Firstly, we compute the sample complexity of each method. Given the number of nodes $|V|$, average degree $\bar{d}$, walk length $L$, window size $W$ and walk time $T$ in RW, and max order $O$ and sampling time $N$ in PS, 
the number of sample pairs under RW and PS are $(2L-W-1)WT|V|$ and $NO\bar{d}|V|$, and the ratio of them $r$ is $\frac{(2L-W-1)WT}{NO\bar{d}}$.
We compute the ratio of node pairs with fine-tuned parameters in Tab. \ref{Tab-Ratios of Different Datasets}.
Note that real ratios will be larger for that the number of pairs sampled by PS is actually no greater than $NO\bar{d}|V|$.
From results in Tab.~\ref{Tab-Ratios of Different Datasets}, we can see that
PS can significantly reduce the training samples (reducing $\sim$99\% samples) compared with RW. 
Note that $r$ is often much larger in sparse networks like Cora, Zhihu and SNOMED.
In graphs with small average degree, RW often conducts DFS-like walks and encounter leaf nodes, which results in frequent revisiting behaviors and hence more redundant samples.

\begin{table}[htbp]
	\centering
	\vspace{-0.3cm}
	\caption{Ratios of Different Datasets}
	\vspace{-0.3cm}
	\begin{tabular}{c|ccc|c}
		\hline
		\textbf{Datasets} & $|V|$ & $|E|$ & $\bar{d}$ & r \\
		\hline
		Cora$@50\%$ & 2,211 & 2,607 & 2.33 & 183.60 \\
		Cora$@100\%$ & 2,211 & 5,214 & 4.42 & 140.46 \\
		Facebook$@50\%$ & 4,039 & 44,117 & 21.84 & 36.17 \\
		Zhihu$@50\%$ & 10,000 & 21,947 & 4.28 & 369.16 \\
		AstroPh$@50\%$ & 18,772 & 99,054 & 10.56 & 55.40 \\
		HepTh$@50\%$ & 27,400 & 176,271 & 12.86 & 99.79 \\
		SNOMED$@20\%$ &391,892 & 409,550 & 2.09 & 409.36 \\
		\hline
	\end{tabular}
	\label{Tab-Ratios of Different Datasets}
	\vspace{-0.3cm}
\end{table}

\subsubsection{Link Prediction}\label{Sec-Link Prediction 1}

We evaluate PS against RW on 6 datasets listed in Section \ref{Sec-Datasets}.
We use PS and RW to generate training sets, and train node embeddings respectively. 
Results in Tab. \ref{Tab-Link Prediction 1} show that 
PS outperforms RW on almost all datasets. 
This indicate that even if 
PS adopts stronger assumptions for efficiency consideration, 
it still proves to be a competitive alternative to RW.

\begin{table}[htbp]
	\centering
	\vspace{-0.3cm}
	\caption{Link Prediction Results of RW and PS}
	\vspace{-0.3cm}
	\begin{tabu}	to 0.45\textwidth{X[2.3,c]|X[c]X[c]X[c]X[c]}
		
		\hline
		\centering \multirow{2}{*}{\textbf{Datasets}} 
		& \multicolumn{2}{c}{\textbf{Random Walk}} & \multicolumn{2}{c}{\textbf{Pairs Sampling}} \\ &
		\multicolumn{1}{c}{$\text{AUC}_\text{LR}$} &\multicolumn{1}{c}{$\text{AUC}_\text{pair}$} &\multicolumn{1}{c}{$\text{AUC}_\text{LR}$} &\multicolumn{1}{c}{$\text{AUC}_\text{pair}$}\\
		\hline
		\centering Cora$@50\%$ & 0.9200 & 0.9293 & \textbf{0.9272} & \textbf{0.9394} \\
		\centering Facebook$@50\%$ & 0.9921 & 0.9892 & \textbf{0.9922} & \textbf{0.9911} \\
		\centering Zhihu$@50\%$ & 0.8659 & \textbf{0.9144} & \textbf{0.8673} & 0.9136 \\
		\centering AstroPh$@50\%$ & 0.9788 & 0.9768 & \textbf{0.9795} & \textbf{0.9789} \\
		\centering HepTh$@50\%$ & 0.9741 & 0.9648 & \textbf{0.9743} & \textbf{0.9730} \\
		\centering SNOMED$@20\%$ & 0.9350 & 0.9396 & \textbf{0.9359} & \textbf{0.9402} \\
		\hline
	\end{tabu}
	\label{Tab-Link Prediction 1}
	\vspace{-0.3cm}
\end{table}

\subsection{Experiments of TGE-PS} 
\subsubsection{Link Prediction}\label{Sec-Link Prediction 2}

We evaluate TGE-PS against all baseline models on SNOMED and HepTh respectively. 
The results are shown in Tab. \ref{Tab-Link Prediction 2}, where ``-'' refers to results of failed experiments.
From Tab. \ref{Tab-Link Prediction 2}, we have following observations:
\begin{itemize}
	\item TGE-PS outperforms all baseline models on both datasets. 
	Incorporating textual information with structural information helps constructing expressive graph embeddings. 
	Improvements of TGE-PS over PS and Paper2Vec over node2vec strongly support this point.
	Only preserving structural information like LINE or textual information like Text Matching both presents limitations.
	\item 
	The improvements of TGE-PS are quite different on the two graphs. 
	Recall that we use the text description of concepts in SNOMED and the whole abstract in HepTh, we owe this to that SNOMED has stronger text dependency than HepTh. 
	The performance of Text Matching in SNOMED supports this point. 
	Similar difference of improvements also appears between Paper2Vec and node2vec.
\end{itemize}

\begin{table}[h]
	\centering
	\vspace{-0.3cm}
	\caption{Link Prediction Results of Different Models}
	\vspace{-0.3cm}
	\begin{tabu}
		to 0.45\textwidth{X[2.3,c]|X[c]X[c]X[c]X[c]}
		\hline
		\centering \multirow{2}{*}{\textbf{Model}} 
		& \multicolumn{2}{c}{\textbf{SNOMED$@20\%$}} &  \multicolumn{2}{c}{\textbf{HepTh$@50\%$}}\\
		
		&\multicolumn{1}{c}{$\text{AUC}_\text{LR}$} &\multicolumn{1}{c}{$\text{AUC}_\text{pair}$} &\multicolumn{1}{c}{$\text{AUC}_\text{LR}$} &\multicolumn{1}{c}{$\text{AUC}_\text{pair}$}\\ 
		\hline
		\centering LINE & 0.6461 & 0.6985 & 0.7883 & 0.7520 \\
		\centering DeepWalk & 0.9164 & 0.9258 & 0.9711 & 0.9573 \\
		\centering node2vec & 0.9350 & 0.9396 & 0.9741 & 0.9648 \\
		\hline
		\centering Text Matching & 0.8954 & 0.8717 & 0.7057 & 0.5700 \\
		\centering TADW & - & - & 0.8866 & 0.8977 \\
		\centering CANE & 0.9613 & 0.9544  & 0.9785 & 0.9388 \\
		\centering Paper2Vec & 0.9581 & 0.9604 & 0.9745 & 0.9748 \\
		\centering STNE & - & - & 0.9651 & 0.9572 \\
		\hline
		\centering PS & 0.9359 & 0.9402 & 0.9758 & 0.9716 \\
		\centering TGE-PS & \textbf{0.9721} & \textbf{0.9621} & \textbf{0.9793} & \textbf{0.9752} \\
		\hline
	\end{tabu}
	\vspace{-0.3cm}
	\label{Tab-Link Prediction 2}
\end{table}

It is also worth noting that not all models are capable of dealing with large graphs:
TADW requires pre-loading the whole adjacent matrix with $O(|V|^2)$ complexity;
the training speed of STNE is extremely slow on astronomical number of walks generated in SNOMED.
We also try to conduct experiments on GraphGAN in link prediction task, but they fail on both datasets due to OOM.  
On the contrary, our TGE-PS performs well in handling large graphs.

\subsubsection{Zero-Shot Experiments} \label{Sec-Zero-Shot Experiments}

Unlike common link prediction scenarios, where each node embedding is trained at least once before inference, 
zero-shot scenarios require prediction on link existence with unseen nodes. 
Zero-shot scenarios are common in real-world applications.
For example, when updating medical terminology graphs like SNOMED, terms of newly found diseases will be added into existing databases,
and it is possible for a human to infer link existence by text information with high accuracy even if he is not an expert in this field.
Therefore, zero-shot scenario in fact reflects whether the model incorporates text information by capturing important message and potential connections or simply matching literal similarities.

We conduct zero-shot experiments on SNOMED and HepTh where 0.5\% of the nodes and related edges are removed from the graph.
Embeddings of these nodes will be generated by trained models directly and used for evaluation.  
As we stated in previous sections, zero-shot scenarios are not widely studied and most current models are incapable of generating embeddings for unseen nodes.
However, to better study this problem, we manage to conduct experiments on a variant of CANE
 with only averaged TE part
 , which we denote as CANE (TE):
We also conduct ablation experiments on zero-shot scenarios to analyze the influence of character- and word-level embeddings. 
In ablation experiments, we either replace $[\Be^{w}; \Be^{w'}]$ with $\Be^{w'}$ (w/o word), or replace $[{\Be^{w}}; \Be^{w'}]$ with $\Be^{w}$ (w/o char).
High AUC scores in Tab. \ref{Tab-Zero-Shot Experiments} indicate TGE-PS is practical and reliable in zero-shot scenarios.
Besides, we have following observations:

\begin{itemize}
	\item Word-level embeddings are essential in embedding nodes. 
	This is not surprising given that there are much more words   (100,471 in SNOMED and 72,083 in HepTh) than characters   (88 in SNOMED and 59 in HepTh, mostly English characters and numbers).
	Besides, when words are often composed of sub-words as in SNOMED, the improvements of character-level embeddings are apparently higher.
	\item Using Text Matching as a baseline, TGE-PS presents larger promotion on HepTh than SNOMED.
	Unlike terms in SNOMED, texts in HepTh are abstracts of papers, which have longer average length. 
	In this case, BiLSTM shows its advantages over naive matching methods in processing long texts.
\end{itemize}

\begin{table}[htbp]
	\centering\vspace{-0.3cm}
	\caption{Zero-Shot Experiment Results}
	\vspace{-0.3cm}
	\begin{tabu}
		to 0.45\textwidth{X[3,c]|X[c]X[c]X[c]X[c]}
		\hline
		\centering \multirow{2}{*}{\textbf{Model}} 
		& \multicolumn{2}{c}{\textbf{SNOMED}} &  \multicolumn{2}{c}{\textbf{HepTh}} \\ &
		\multicolumn{1}{c}{$\text{AUC}_\text{LR}$} &\multicolumn{1}{c}{$\text{AUC}_\text{pair}$} &\multicolumn{1}{c}{$\text{AUC}_\text{LR}$} &\multicolumn{1}{c}{$\text{AUC}_\text{pair}$}\\ 
		\hline
		\centering Text Matching & 0.9059 & 0.8813 & 0.5934 & 0.6250 \\
		\centering CANE  (TE) & 0.5271 & 0.5344 & 0.6036 & 0.5288 \\
		\hline
		\centering TGE-PS (w/o word) & 0.5000 & 0.5003 & 0.5000 & 0.5037 \\
		\centering TGE-PS (w/o char) & 0.9701 & 0.9786 & 0.8979 & \textbf{0.9485} \\
		\centering TGE-PS & \textbf{0.9760} & \textbf{0.9811} & \textbf{0.8990} & \textbf{0.9485} \\
		\hline
	\end{tabu}
	\label{Tab-Zero-Shot Experiments}
	\vspace{-0.3cm}
\end{table}

\section*{Acknowledgments}
The work is supported by National Natural Science Foundation of China (61702327, 61772333, 81771937), Shanghai Sailing Program (17YF1428200).
Weinan Zhang and Yong Yu are the corresponding authors.
\newpage
\bibliographystyle{ACM-Reference-Format}
\balance 
\bibliography{sample-bibliography}

\newpage
\appendix
\section{Pseudo-code}
The pseudo-code for PS is given in Algorithm \ref{Alg-Pairs Sampling}. 

\begin{algorithm}[htbp]
	\begin{algorithmic}
		\STATE {\bf LearnFeatures}{(Graph $G=(V, E)$, Dimensions $dim$, Max order $O$, Sampling time $N$)}
		\STATE Initialize pairs set $\mathcal{P}$ to Empty\\
		\FOR {$n = 1$ {\bfseries to} $N$} 
		\FORALL {nodes $u \in V$}
		\STATE Initialize previous neighbors $\mathcal{N}_\text{prev}$ and pairs $\mathcal{P}'$ to Empty
		\FOR {$o = 1$ {\bfseries to} $O$} 
		\STATE $\mathcal{P}', \mathcal{N}_\text{prev} =$ Sampling($G, u, o, \mathcal{N}_\text{prev}$)
		\STATE 
		$\mathcal{P}\gets\mathcal{P}\cup\mathcal{P}'$
		\ENDFOR
		\ENDFOR
		\ENDFOR
		\STATE $f =$ AdamOptimizer($d$, $\mathcal{P}$)
		\STATE {\bf return} $f$
		\STATE \hrulefill
		\STATE {\bf Sampling}{ (Graph $G= (V, E)$, Start node $u$, Order $o$, Previous neighbors $\mathcal{N}_\text{prev}$)}
		\STATE Inititalize neighbors $\mathcal{N}$ to Empty
		\STATE Inititalize pairs $\mathcal{P}'$ to Empty
		\FORALL {$w \in \mathcal{N}_\text{prev}$}
		\FORALL {$v \in$ GetNeighbors($G, w$)}
		\IF {distance($u, v$) = $o$}
		\STATE
		$\mathcal{N}\gets\mathcal{N}\cup\{v\}$
		
		\STATE \bfseries{break}
		\ENDIF
		\ENDFOR
		\STATE 
		$\mathcal{P}'\gets \mathcal{P}'\cup\{(u, v)\}$
		\ENDFOR
		\STATE {\bf return} $\mathcal{P}', \mathcal{N}$
	\end{algorithmic}
	\caption{The Pairs Sampling Algorithm.}
	\label{Alg-Pairs Sampling}
\end{algorithm} 
\end{document}